\pdfoutput=1
\documentclass[letterpaper, 10 pt, conference]{ieeeconf}

\IEEEoverridecommandlockouts
\usepackage[dvipdfmx]{graphicx} \usepackage{bmpsize}
\usepackage[pagebackref=true]{hyperref}
\usepackage{amsmath}
\usepackage{tabularx,amsmath,amsfonts,siunitx,pifont,amssymb,caption,svg}
\usepackage{xcolor, bm, float, multirow, courier, mathtools, url}
\usepackage[ruled,vlined]{algorithm2e} 

\let\labelindent\relax \usepackage{enumitem}

\usepackage{scalerel}
\usepackage{tikz}
\usetikzlibrary{svg.path}

\definecolor{orcidlogocol}{HTML}{A6CE39}
\tikzset{
	orcidlogo/.pic={
		\fill[orcidlogocol] svg{M256,128c0,70.7-57.3,128-128,128C57.3,256,0,198.7,0,128C0,57.3,57.3,0,128,0C198.7,0,256,57.3,256,128z};
		\fill[white] svg{M86.3,186.2H70.9V79.1h15.4v48.4V186.2z}
		svg{M108.9,79.1h41.6c39.6,0,57,28.3,57,53.6c0,27.5-21.5,53.6-56.8,53.6h-41.8V79.1z M124.3,172.4h24.5c34.9,0,42.9-26.5,42.9-39.7c0-21.5-13.7-39.7-43.7-39.7h-23.7V172.4z}
		svg{M88.7,56.8c0,5.5-4.5,10.1-10.1,10.1c-5.6,0-10.1-4.6-10.1-10.1c0-5.6,4.5-10.1,10.1-10.1C84.2,46.7,88.7,51.3,88.7,56.8z};
	}
}

\newcommand\orcidicon[1]{\href{https://orcid.org/#1}{\mbox{\scalerel*{
				\begin{tikzpicture}[yscale=-1,transform shape]
				\pic{orcidlogo};
				\end{tikzpicture}
			}{|}}}}

\newcommand{\ic}{\mathbf x}
\newcommand{\icset}{X}
\newcommand{\icspace}{\mathcal X}
\newcommand{\opti}{\mathbf y}
\newcommand{\optiset}{Y}
\newcommand{\optispace}{\mathcal Y}
\newcommand{\learnedparams}{\theta}
\newcommand{\cost}{g}

\newcommand{\heuristic}{\emph{Heuristic}}
\newcommand{\learned}{\emph{LearnedInit}}

\usepackage{pdfpages}

\makeatletter
\let\NAT@parse\undefined
\makeatother

\title{\LARGE \bf
   Reliable Trajectories for Dynamic Quadrupeds \\
   using Analytical Costs and Learned Initializations} 

\author{Oliwier Melon \orcidicon{0000-0001-6092-9477}, 
		Mathieu Geisert \orcidicon{0000-0002-5651-8736}, 
		David Surovik \orcidicon{0000-0002-9454-5874}, 
		Ioannis Havoutis \orcidicon{0000-0002-4371-4623} 
		and Maurice Fallon \orcidicon{0000-0003-2940-0879}
\thanks{This work was supported by the UKRI/EPSRC RAIN Hub
[EP/R026084/1] and the
EU H2020 Projects MEMMO and THING, the EPSRC grant `Robust Legged Locomotion' [EP/S002383/1] and a Royal Society University Research Fellowship (Fallon). This work was conducted as part of ANYmal Research, a community to advance legged robotics.
The authors are with Oxford Robotics
Institute, University of Oxford, UK. Email:
{\tt\small \{omelon, mathieu, dsurovik, ioannis, mfallon\}@robots.ox.ac.uk}.}	
}

\makeatletter
\def\endthegraphy{%
    \def\@noitemerr{\@latex@warning{Empty `thebibliography' environment}}%
    \endlist
}
\makeatother

\begin{document}
	
\setlength{\abovedisplayskip}{4pt}
\setlength{\belowdisplayskip}{4pt}
	
\maketitle 
\thispagestyle{empty} 
\pagestyle{empty}
	
\begin{abstract}
Dynamic traversal of uneven terrain is a major objective in the field of legged robotics.
The most recent model predictive control approaches for these systems can generate robust dynamic motion of short duration;
however, planning over a longer time horizon may be necessary when navigating complex terrain.
A recently-developed framework, Trajectory Optimization for Walking Robots (TOWR), computes such plans but does not guarantee their reliability on real platforms, under uncertainty and perturbations.
We extend TOWR with analytical costs to generate trajectories that a state-of-the-art whole-body tracking controller can successfully execute.
To reduce online computation time, we implement a learning-based scheme for initialization of the nonlinear program based on offline experience.
The execution of trajectories as long as 16 footsteps and 5.5~s over different terrains by a real quadruped demonstrates the effectiveness of the approach on hardware.
This work builds toward an online system which can efficiently and robustly replan dynamic trajectories.

\end{abstract}
	
\section{Introduction} \label{sec:intro}

Legged robots are soon expected to be used in a range of application domains where advanced mobility is required. 
The key benefit of such machines is their flexibility in operating on terrain designed
for humans: confined spaces and lacking regular structure.
This goal implies increased complexity as legged robots are articulated systems with high-dimensional kinematics and dynamics that continually change their contact with the environment.
Other hurdles arise due to real-world sensing, actuation limits, state estimation and perturbations.
As a result, legged robots need a flexible motion planning approach to efficiently and robustly perform their tasks. 

The generation of dynamic motions for these platforms is an open research problem with recent advances focusing on optimization-based approaches.
Longer trajectories can produce better system performance, but are more difficult to compute, ruling out the online use of global methods.
Gradient descent is less demanding but can get stuck in poor local minima due to the strong non-convexity of the problem.
Thus, local optimization is appealing only if the constraints, costs, and the initial guess of the nonlinear program can all be specified effectively.

\begin{figure}[t]
	\includegraphics[width=\columnwidth]{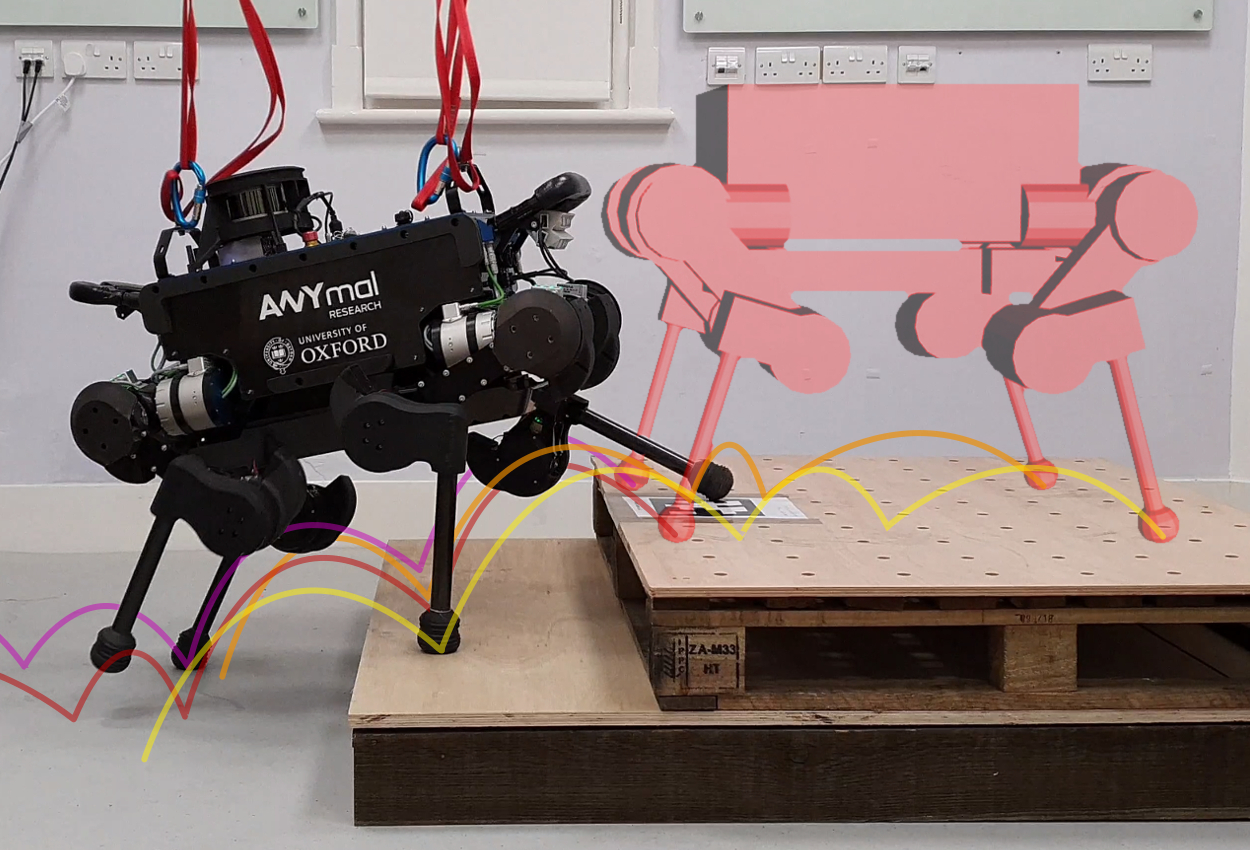}
	\caption{The ANYbotics ANYmal executing a dynamic stair climb with the proposed approach, as shown in the accompanying video \small{\url{https://youtu.be/LKFDB_BOhl0}}.}
	\label{fig:experiment}
\end{figure}

\begin{figure}
	\includegraphics[width=\columnwidth]{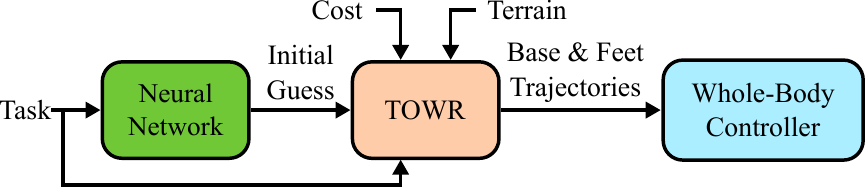}
	\caption{Overview of the system used for learning, trajectory optimization and execution on the ANYmal quadruped.}
	\label{fig:system_overview}
	\vspace{-3mm}
\end{figure}

As shown in Fig.~\ref{fig:experiment} and \ref{fig:system_overview}, we combine a learning-based data-driven initialization with an enhanced formulation of the optimization problem of dynamic motion generation, which we demonstrate for a quadrupedal robot walking on uneven terrain.
The objective is to produce robust plans over suitably long time horizons with minimal online computation effort.


\subsection{Contributions}
\begin{itemize}
\item Extension of the TOWR legged robot motion optimization framework \cite{winklerGaitTrajectoryOptimization2018} with analytical costs to create dynamic trajectories that can be reliably tracked by a controller.
\item A data-driven method to produce good initial guesses which speeds up optimization convergence. The method can compute 16-footstep trajectories in less than 1 second while avoiding poor local minima.
\item Significant evaluation in dynamic simulation to demonstrate how the combination of these changes make execution much more reliable than the baseline method.
\item Experiments in which the real ANYmal robot trots or walks across flat or uneven terrain while using only onboard sensors for state estimation, verifying the validity of our approach and its suitability for real hardware.
\end{itemize}

\subsection{Overview}
This paper begins with an overview of related state-of-the-art approaches in Sec.~\ref{sec:related_work}.
Descriptions of the baseline trajectory optimizer and tracking controller are then provided in Sec.~\ref{sec:loco}, along with new modifications that allow these tools to be effectively combined.
Sec.~\ref{sec:learning_method} describes our method for training effective initialization using offline prior experiences, with some analysis of its benefit.
Sec.~\ref{sec:expe} describes experimentation on different terrains in both simulation and hardware.
The paper concludes with final remarks in Sec.~\ref{sec:conclu}.

\section{Related Work}\label{sec:related_work}
\subsubsection{Trajectory Optimization for Legged Robots}
Trajectory optimization (TO) approaches have been used for short to medium scale motion planning in recent research~\cite{diehlFastDirectMultiple2006,posaDirectMethodTrajectory2014}.
They usually employ direct methods to transcribe a continuous, infinite-dimensional problem as a discrete, parametrized nonlinear programming problem (NLP) \cite{bettsPracticalMethodsOptimal2010}.

Legged locomotion optimization problems are subject to discontinuities due to contact transitions of the end-effectors.
Two approaches have been used to recast it as a continuous problem. 
The first reformulates contact as a smooth rather than a discontinuous state \cite{TodorovConvexSmoothInvertible2011, erezTrajectoryOptimizationDomains2012,mordatchDiscoveryComplexBehaviors2012}.
This transforms the discontinuous and minima-prone problem into a continuous problem that can be solved with homotopic methods.

The second approach defines the problem as a succession of phases separated by the contact state transition of each end-effector. 
In most cases, the timing and position of the contacts are computed externally and the optimization only solves for the centroidal motion of robot \cite{pontonConvexModelHumanoid2016,carpentierVersatileEfficientPattern2016a,icra2017mastalli,bellicosoDynamicLocomotionOnline2018}.
In \cite{winklerGaitTrajectoryOptimization2018}, the contact state of each leg is considered separately, which theoretically allows it to generate different gaits when optimizing the duration of each phase.
Moreover, the positions of the footsteps are included in the set of optimized variables.
Unlike the previously mentioned approaches, this formulation allows a highly non-convex shape of the problem to be solved by recasting it as a feasibility problem. 

\subsubsection{Data-driven Initialization} \label{sec:rw_learning}
Data-driven trajectory initialization schemes have been applied in the domain of manipulation to speed up the computation of smooth paths when reaching past obstacles~\cite{berensonRobotPathPlanning2012,jetchevFastMotionPlanning2013,merktLeveragingPrecomputationProblem2018}. 
Related methods produce multiple initializations in different basins of attraction so as to identify distinct ways of approaching the object to be grasped~\cite{draganLearningFromExperience2011,lembonoMemoryOfMotion2019}.
Dynamic constraints can additionally be met for tasks 
such as quickly reaching to catch a thrown object~\cite{lamparielloTrajectoryPlanningOptimal2011} or rejecting large disturbances on underactuated aerial vehicles~\cite{mansardUsingMemoryMotion2018}.
For legged locomotion on terrain, a related idea was used to plan individual feasible footsteps, which were then combined by another process into a full motion plan~\cite{stolleTransferOfPolicies2007}.

Most of these efforts map tasks to initial solutions, while others map to segments of solutions~\cite{stolleTransferOfPolicies2007}, or to additional constraints that convexify the problem~\cite{draganLearningFromExperience2011}.
Nearly all consider nearest-neighbor lookup and regression on an experience library as a mapping method, while many also consider other function approximators.
These include
Support Vector Machines~\cite{draganLearningFromExperience2011, lamparielloTrajectoryPlanningOptimal2011}, 
Gaussian Process Regression~\cite{lembonoMemoryOfMotion2019,lamparielloTrajectoryPlanningOptimal2011}, 
or Artificial Neural Networks (ANNs)~\cite{draganLearningFromExperience2011,mansardUsingMemoryMotion2018}. 
Another research strand is the use of feature-spaces and dimensionality reduction techniques~\cite{stolleTransferOfPolicies2007,jetchevFastMotionPlanning2013,merktLeveragingPrecomputationProblem2018,draganLearningFromExperience2011,lembonoMemoryOfMotion2019}.
In each case, the goal is to sufficiently represent past experiences in a manner which can be related to future decisions. 
\section{Motion Generation} \label{sec:loco}

In this section we present our motion planning and execution approach.
We first review the TOWR trajectory optimization package, which uses a simplified dynamics model of the robot to plan motions for its legs and center of mass (CoM) between initial and final configurations.
We present improvements and adaptions that help create trajectories which are more suitable for the real robot.
We then overview the tracking whole-body controller that we use to compute the joint torques necessary to execute these generated motions and provide an illustrative evaluation.

\subsection{Trajectory Optimization for Walking Robots (TOWR)}
\label{ssec:towr}
The paper extends the work of Winkler \emph{et~al.} and their open-source library TOWR\cite{winklerGaitTrajectoryOptimization2018,winklerTOWROpensourceTrajecetory2018}.
TOWR is capable of producing highly-dynamic trajectories for a range of walking robots by
formulating locomotion as a nonlinear program (NLP). 
The approach considers single rigid body dynamics (SRBD) of the base, 
which is assumed to contain all of the system's mass, along with the paths and contact forces of the feet. 

\begin{figure}[tb]
	\centering
	\includegraphics[width=\columnwidth]{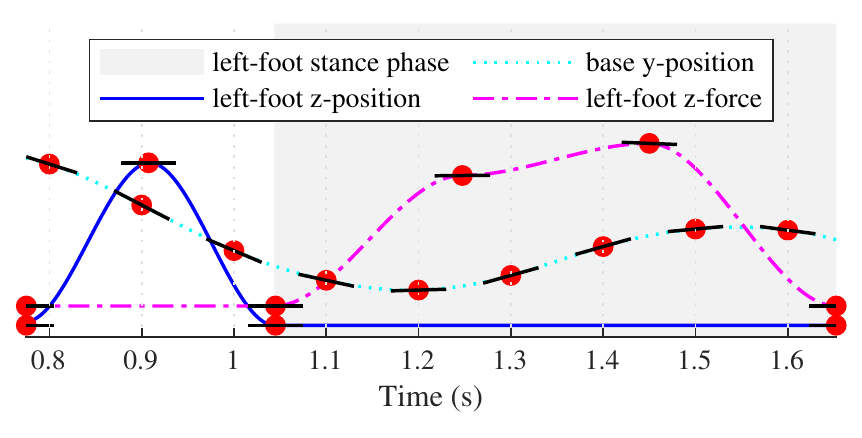}
	\caption{
		An illustrative sample of the trajectories computed by the numerical optimizer, for a quadrupedal trot. Red dots and black lines represent values at optimization nodes, interpolated using cubic splines. 
	}
	\label{fig:towr_splines}
	\vspace{-5mm}
\end{figure}

The problem is discretized into a numerically-solvable formulation using a collocation-based transcription method. 
In this case, trajectories are constructed as splines of $N$ cubic Hermite polynomials, where each polynomial is fully defined by the values and the derivative at its start and end nodes.
Fig.~\ref{fig:towr_splines} shows these splines for some of the problem variables.
The base trajectory is discretized using a fixed timestep $dt$ (0.1~s), while the feet trajectories and contact forces are discretized with a fixed number of polynomials per phase (2 for the swing trajectory, 3 for the stance forces).
Therefore, the number of variables for the feet and contact forces varies with the number of steps, while the number of base-related variables depends on the time horizon.
The formulation of the locomotion problem implicitly constrains some of the variables: 
\begin{itemize}
\item Forces are null during swing phases.
\item Derivatives of the forces and feet positions are zero at the transitions between swing and stance phases.
\item A swing node is the highest point of a swing trajectory and its z-dimension derivative is set to 0.
\item Feet are fixed in place during a stance phase. 
\end{itemize}
Consequently, the number of optimization variables is then $N = 12T/dt + 20S + 120$,
where $T$ is the time horizon, $dt$ the time discretization interval for the base motion and $S$ is the number of steps.
Moreover, if the timing of steps is optimized, the duration of the stance and swing phases of each step are included, adding $2S$ variables.

The problem is then transferred to an interior-point method solver (Ipopt \cite{wachterImplementationInteriorpointFilter2006}) which searches for a solution respecting the following explicit constraints:
\begin{itemize}
	\item Dynamics of the system (modeled as one rigid body).
	\item Kinematic limits (the positions of feet, relative to the base, constrained within a box).
	\item Maximum contact forces and friction pyramids.
	\item Feet at terrain height during the stance phase.
\end{itemize}
Since the original TOWR formulation poses only a feasibility problem, the NLP solver generally converges quickly (in less than 40 iterations).
While the resulting motion plans are promising, realizing them on hardware is difficult.
Oscillation and strong body rolling motions are common while footstep placement and leg terrain clearance are not considered. 
In this work, we tackle these issues as follows.

\subsubsection{Smoothing trajectories using costs}
To generate more conservative trajectories for motions such as the one shown in Fig. \ref{fig:experiment}, we extended TOWR with analytically-derived costs.
A set of $i$ costs $\mathrm{J}_{i}(t)$ are scaled by weights $\omega_{i}$, to give a total cost of $\mathrm{J}(t) = \sum \omega_{i} \mathrm{J}_i(t)$.
Integral costs were added to minimize linear velocity in the z-axis and angular velocities of the base,
to penalize the magnitude of the ground-reaction forces and their derivatives, 
and to maintain the desired magnitude of the normal force.
The generic cost was
\begin{equation} \label{eq:integral_cost}
	\mathrm{J}_{i}(t) = \int_{0}^{T} \left ( \left ( x_i(t) - x^{ref}_i(t) \right ) ^2 + \omega_{i,d}{\dot{x}_i(t)}^2 \right ) dt,
\end{equation}
where $x_i(t)$ is the optimized polynomial and $\omega_{i,d}$ is the weight on its derivative.
This penalizes short, but potentially large, deviations from the reference value $x^{ref}_i(t)$.
These costs and their Jacobians were computed analytically using the parameters of the polynomials and their relations to the values at each node of the trajectories.

\subsubsection{Locomotion on uneven terrain}
We aim to generate dynamic motions to traverse slopes, steps and stairs. The original TOWR implementation does not take into account swing collisions between spline nodes.
To resolve this, we add kinematic constraints to the base of the robot to enforce that it remains a certain distance above the feet; this eliminates the possibility of collision between the base and the ground.
To prevent the solver from creating trajectories that pass through or collide with step edges,
we added a cost to discourage the robot from selecting footsteps close to these step edges.
The selected cost is a differentiable Gaussian function $\sum e^{x_s^2/2\sigma^2}$ where $x_s$ represents the perpendicular distance from each footstep to the edge of the stair.
This only affects the footsteps which are close to the edge, while having a negligible effect on the remaining ones.
To ensure that collisions with terrain are avoided, both a constraint and a cost are applied to the height of the swinging feet. The constraint ensures that the swing node is a certain distance above each of the adjacent stance nodes while
the cost minimizes the swing height to prevent large leg motions that would create angular momentum on the real system.

\subsection{Whole-Body Controller}
\label{ssec:wbc}
Once the trajectory has been generated, we use the whole-body controller of Bellicoso \emph{et al.}~\cite{bellicosoPerceptionlessTerrainAdaptation2016}  to track the trajectory of the base and the end-effectors at \SI{400}{\hertz}. The controller contains a state machine which adapts gains in response to slipping and other unexpected contact events. 
\subsection{Validation in Simulation}
\begin{figure}[t]
	\centering
	\includegraphics[width=\columnwidth]{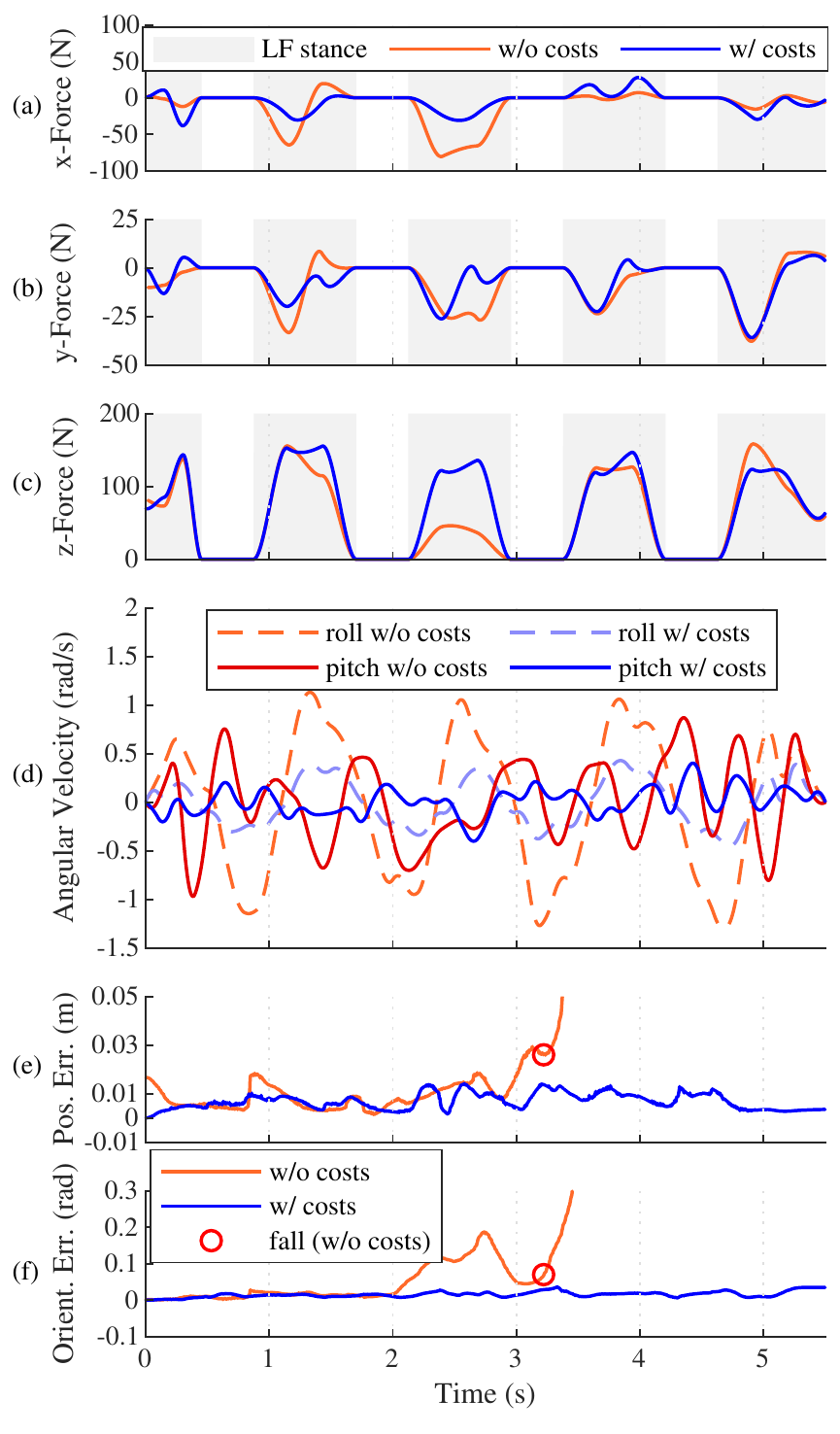}
	\caption{
	    Simulated climbing a single step, similar to Fig.~\ref{fig:pallet_snpashots}:
		by adding costs to the optimization problem we can constrain foot forces and base angular velocities to successfully climb a step (Sec.~\ref{ssec:single-pallet}) in simulation.}
	\label{fig:costs_effects}
\end{figure}

In Fig.~\ref{fig:costs_effects}, we demonstrate the need for more sophisticated planning for a trajectory up and over a step, both with and without the proposed costs.
Fig.~\ref{fig:costs_effects}(a) and \ref{fig:costs_effects}(b) show an overall decrease in peak x- and y-direction forces.
Fig.~\ref{fig:costs_effects}(c) shows an increase in the z-direction force during the third second of the trajectory resulting from a corresponding decrease in the x-direction force.
Fig.~\ref{fig:costs_effects}(d) shows the inertial stabilization of the base---a significant reduction in the roll and pitch angular velocity of the base can be observed.

Fig.~\ref{fig:costs_effects}(e) and \ref{fig:costs_effects}(f) show the norm of the tracking errors of the base during the execution of the trajectories by the whole-body controller in the Gazebo physics simulator.
Initially, the controller manages to track both trajectories well; after 2~s, the controller can no longer adequately follow the trajectory without the proposed costs, as indicated by the increase in orientation error.
At 3.2~s, the robot's front foot collides with the step and the robot falls.
Meanwhile, the controller tracking the trajectory generated with the addition of the proposed costs successfully completes the execution.

However, the use of costs and constraints makes the solve time of the optimization problem longer, primarily by increasing the number of iterations; we use machine learning to provide an efficient optimization seed which offsets the extra computation time shown in Table~\ref{tab:computation_time}.
\section{Learning Initializations} \label{sec:learning_method}
Optimization frameworks such as TOWR use solvers which require a guess of the solution $\opti$. It can be generated automatically by some map 
\begin{align} 
A:\ic\rightarrow\opti_0
\end{align}
that acts upon the task $\ic$, e.g., the pair of initial and desired robot states. 
Here, primal variables of the interior-point method are initialized with a guess which substantially affects not only the rate of convergence but also the quality of $\opti$.
As more costs and constraints are used, more local minima arise, and the initial guess becomes even more influential.

The conventional guess generator for TOWR, termed \emph{Heuristic}, linearly interpolates a path for the floating base between the start and goal locations from $\ic$.
Footsteps are evenly-spaced and transitions between them are evenly-timed according to the selected gait, with contact forces equally distributed to counteract the robot's weight.

\begin{figure}[!t]
    \centering
	\includegraphics[width=\columnwidth]{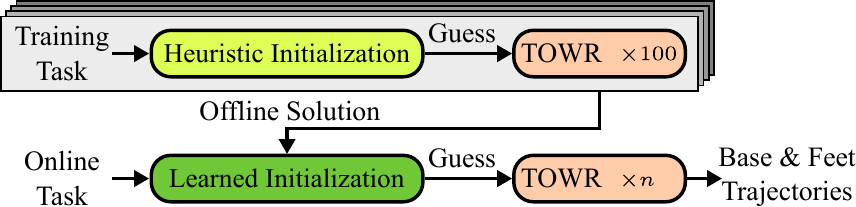}
	\caption{Learning from previous fully-converged outcomes allows the optimizer to be initialized close to a good solution.}
	\label{fig:learning-diagram}
\end{figure}

The objective of this section is to produce a data-driven initializer, \emph{LearnedInit}, to replace the heuristic such that far less optimization effort (iterations) is required, as sketched in Fig.~\ref{fig:learning-diagram}, while furthermore avoiding poor local minima.

\subsection{Methodology} 
Denoting the heuristic as $A_h$ and an iteration of the optimizer as another map $T$, optimization can be expressed as $\opti_{h,N}\left(\ic\right)=T^NA_h\ic$, with iteration count $N$ selected based upon convergence or available computation time.
The improved initializer is a function approximator $A_\learnedparams$ trained to mimic the behavior of the optimization process $T^NA_h$.
For a given task space $\icspace$, its optimal parameters are then defined by
\begin{align}
    \learnedparams = \underset{\learnedparams^\prime}{\mathrm{argmin}}\underset{\ic\in\icspace}{\sum} \| A_{\learnedparams^\prime}\ic - \opti_{h,N}\left(\ic\right) 
    \|_W \label{eq:learning}
\end{align}
with a positive-definite weight matrix $W$.

In this work $A_\learnedparams$ is a fully-connected neural network with 2 to 3 hidden layers, where $\learnedparams$ refers to the connection weights.
These are determined through supervised learning on a dataset $D = \left\{\ic_i,\opti_{h,N}\left(\ic_i\right)\right\} = \left(\icset, \optiset\right)$ with samples $\ic_i\in\icspace$.
At present, a given $\learnedparams$ is learned for a specific environment; however, this serves as a first step toward contextual planning with $\ic$ augmented by local environmental features.
The form of $\opti$, whose length depends on the total duration and footstep count, is also kept constant.

Given the significant risk of converging to poor local minima, the dataset $D$ is not guaranteed to imply a well-behaved map from $\mathcal X$ to the solution space $\optispace$.
Two additional steps are thus taken to ensure the tractability of the learning problem (\ref{eq:learning}) and the quality of its result.

First, $D$ is filtered based on a threshold of solution cost $\cost_{max}$ to exclude poor solutions from training:
\begin{align}
D_{good} = \left\{\ic_i, \opti_i \;|\; \cost(\opti_i) < \cost_{max}\right\}
\end{align}
Second, since even $D_{good}$ is unlikely to contain only globally optimal solutions, the average performance and uniformity of the learned initialization can potentially be increased by repeating the process of Fig.~\ref{fig:learning-diagram} with $\theta$ retrained on optimization outcomes resulting from its previous value. 
This cycle of moderate exploration and filtering thus lends an aspect of reinforcement learning to the scheme, 
with particular similarity to the alternation between local optimization and global supervised learning of control laws in Guided Policy Search~\cite{levineLearningNeuralNetwork2014}.

The learning method is summarized in Algorithm~\ref{algo:learning}. 
Notably, $N_\learnedparams < N_h$ can be used due to the faster convergence observed when using learned initialization. 

\begin{algorithm}
	\DontPrintSemicolon
	\KwIn{A set of sampled tasks $\icset$}
	\KwOut{Learned initializer parameters $\learnedparams$}
	$\optiset \gets$\sc{Towr}$(A_h(\icset); N_h)$\\
	\For{loop count}{
		$D_{good} \gets$\sc{Filter}$(\icset,\optiset;g_{max})$\\
        $\learnedparams \gets$\sc{SupervisedLearning}$(D_{good})$\\
		$\optiset \gets$\sc{Towr}$(A_{\learnedparams}(\icset); N_\learnedparams)$\\
	}
	\Return{$\learnedparams$}\\
    \caption{{\sc TrainInitializer}}
	\label{algo:learning}
\end{algorithm} 

\subsection{Setup}
For each of the test environments, which will be fully detailed in Sec.~\ref{sec:expe}, a set of about 2000 tasks were sampled from the task space.
This space expresses variation of the initial base location and yaw angle. 
The distribution $\cost\left(\optiset\right)$ of costs resulting from optimization with \heuristic{} and $N_h=100$ generally exhibited a long tail of outliers, as seen in Fig.~\ref{fig:learning-opt}, that correspond to poor local minima.
The filtering threshold $\cost_{max}$ was set at the start of this tail, reducing the initial training set $D_{good}$ to about 300-500 sampled tasks, and a two-layer network was fit to this small dataset.
Subsequent learning cycles used $N_\learnedparams=25$ due to fast convergence, as well as a 3-layer network due to the larger number of samples passing through the filter.

The loss-weighting matrix $W$ consistently weighted all optimization variables within a given category.
An initial check was conducted in which some categories were initialized by $A_\learnedparams$ and others by $A_h$.
This revealed that the base linear positions, stepping phase durations, and contact forces were the most crucial values to improve through learned initialization.
The implication is that it is easier for the optimizer to move along some dimensions of the solution space than others, and that the initializer should prioritize the accuracy of the strongest nonlinear influences.
Ultimately, the best performance was obtained using coarsely tuned weights that included all categories.

\subsection{Analysis}

\begin{figure}[tb]
	\includegraphics[width=\columnwidth]{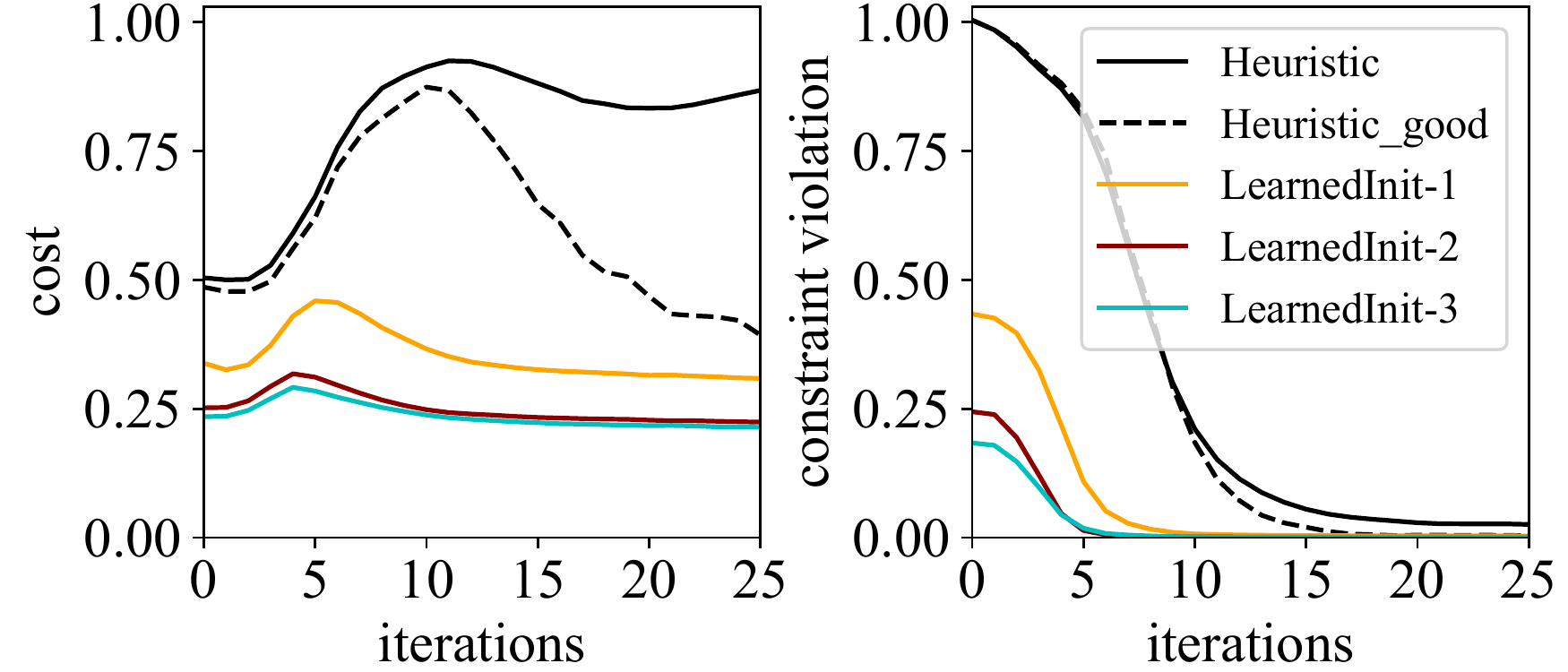}\\
	\vspace{0.25mm} \\
	\includegraphics[width=\columnwidth]{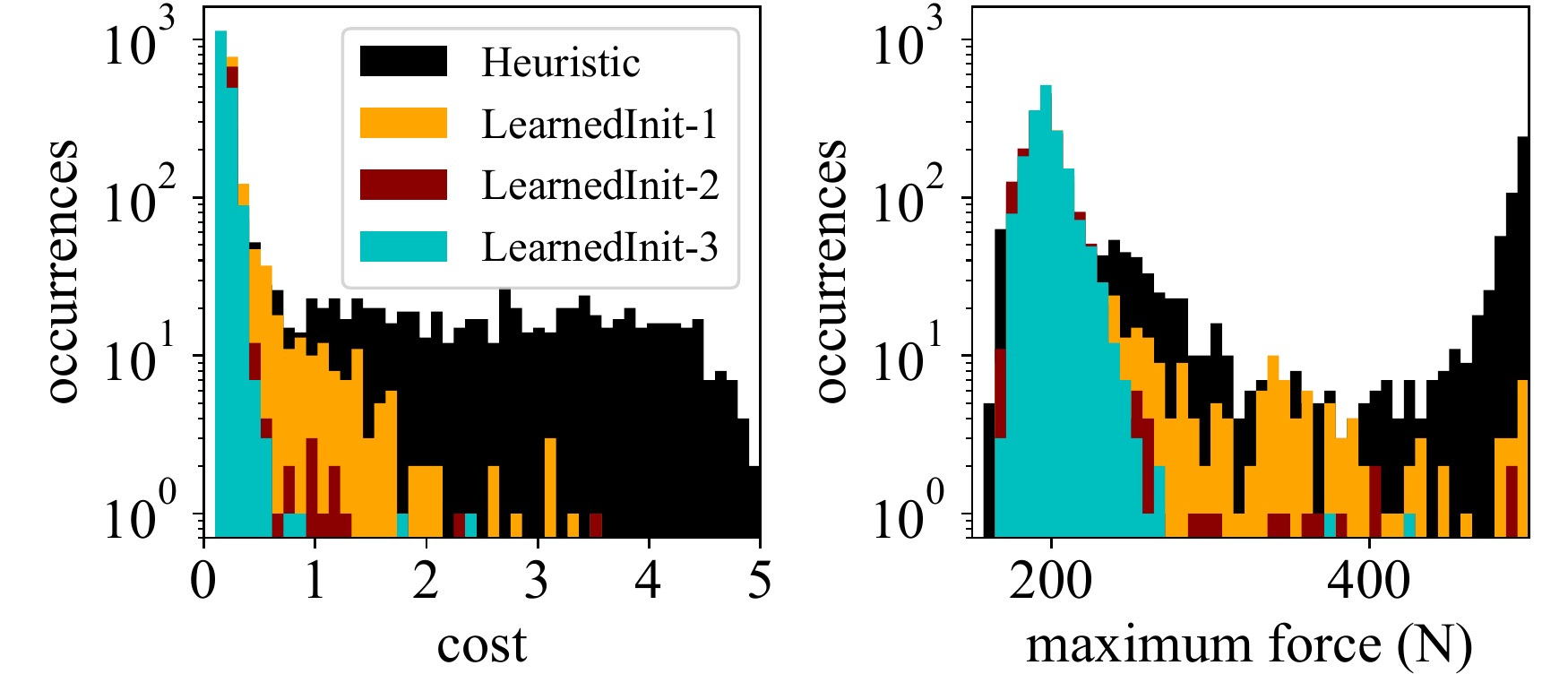}
	\caption{
		Optimization performance under different initialization schemes on a large task set. 
		With \heuristic, many sampled tasks diverge, producing high costs and forces.
		Each cycle of training for \learned{} improves the convergence rate, final cost, and maximal force, while eliminating outliers. 
	}
	\label{fig:learning-opt}
	\vspace{-5mm}
\end{figure}

This section provides a detailed look at the effect of Algorithm~\ref{algo:learning} upon the optimization process.
As similar trends were observed for all the environments tested, their detailed description is deferred to Sec.~\ref{sec:expe}.
The results here reflect the \emph{Single Pallet} problem of ascending a ledge.
For this problem, the starting base position ranges from between $-0.5$~m to $-1.5$~m back from the step and $\pm 0.75$~m laterally from the goal position, with yaw variation in the range $\pm 30^\circ$.

Figure~\ref{fig:learning-opt} shows the benefits of \learned{} in terms of convergence rate, final cost, and the maximum force experienced by the robot.
\heuristic{} initialization often results in poor local minima, and sometimes causes divergence of the optimization process as indicated by high costs and violations of the maximum force constraint (truncated from the plot).  
Due to the use of filtering, \learned{} exhibits better and more consistent performance, succeeding on most tasks that were failed by \heuristic{}.
Retraining \learned{} after optimizing its original output set increases these benefits and eliminates nearly all outliers.

\section{Experimental Evaluation} \label{sec:expe}
\begin{figure*}
	\includegraphics[width=\textwidth]{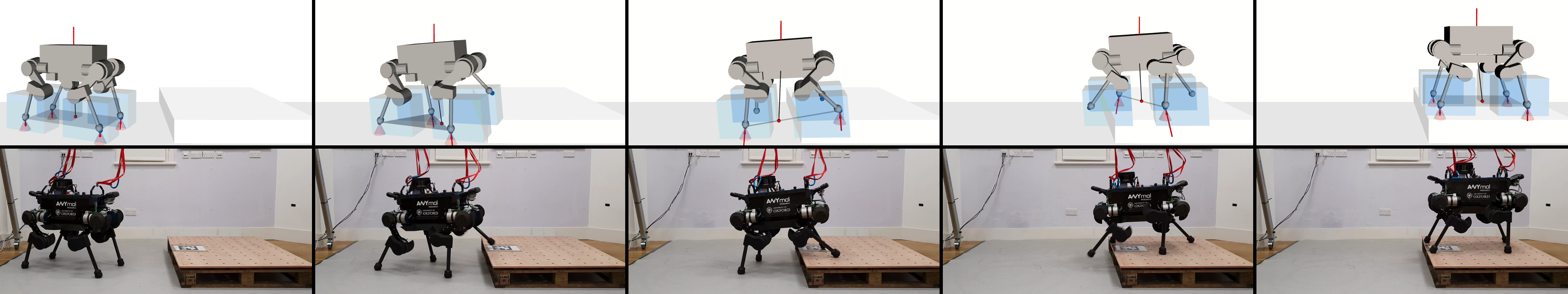}
	\caption{The comparison of a TOWR-generated, dynamic trajectory (top) with the experimental evaluation (bottom) for a 5.5~s climbing of a pallet with a turn.}
	\label{fig:pallet_snpashots}
	\vspace{-5mm}
\end{figure*}

The proposed approach was evaluated using 3 different terrains of increasing difficulty.
As discussed in Sec.~\ref{ssec:towr}, the number of optimization variables depends on the number of steps and the time horizon of the trajectory.
While the number of steps was kept constant (16 steps for dynamic walking, 14 for trotting), the time horizon was changed on a per-environment basis to make dynamic trajectories feasible.
A different initializer was trained for each gait and environment pair, corresponding to a fixed number of optimization variables.

\subsubsection{Flat ground}
The generated trajectories were tested on flat ground for distances of up to 1.5~m with strides of up to 50~cm.
To have highly dynamic motions, the time horizon has been set to 3.5~s which resulted in a velocity of about 0.6~m/s during the middle stage of the trajectory.
The first iteration of the learning phase shown in Fig.~\ref{fig:learning-diagram}, the heuristic initialization, used to generate the first set of data, based the contact sequence on a trotting gait.
For this setup, the number of optimization variables was 848.

\subsubsection{Single Pallet}
\label{ssec:single-pallet}
The Single Pallet is a standard 1.2~$\times$~1~m industrial pallet with a sheet of plywood on top (see Fig.~\ref{fig:pallet_snpashots}), whose total height was 16.5~cm.
For this experiment, the forward motion was about the same as for the Flat Ground, while the time horizon of the trajectory was increased to 5.5~s.
The initial orientation of the base was restricted such that the pallet stays within the field of view of the robot's camera (to detect the pallet position).
In this scenario, there were 952 optimization variables.

\subsubsection{Double Pallet}
This two-step environment is shown in Fig.~\ref{fig:experiment}.
The steps were 14.5~cm and 16.5~cm high.
The horizontal distance between steps was 40~cm. The optimization parameters were kept the same as for the Single Pallet; again being 952.

\subsection{Test in dynamic simulation}
\label{sec:expe:sim_result}

The tracking errors of the desired base pose for the Single Pallet are shown in Fig.~\ref{fig:costs_effects}(e) and \ref{fig:costs_effects}(f).
To evaluate the performance of the learning-based approach, solutions from three initialization methods were tested on the Double Pallet.
While varying the maximum iteration count, a sample set of 100 tasks was optimized for each. 
Trials that did not reach the goal state were marked as failures.
Tracking accuracy was measured for successful runs to indicate how well-suited the planned trajectories were to the closed-loop system.

\begin{figure}[tb]
	\includegraphics[width=\columnwidth]{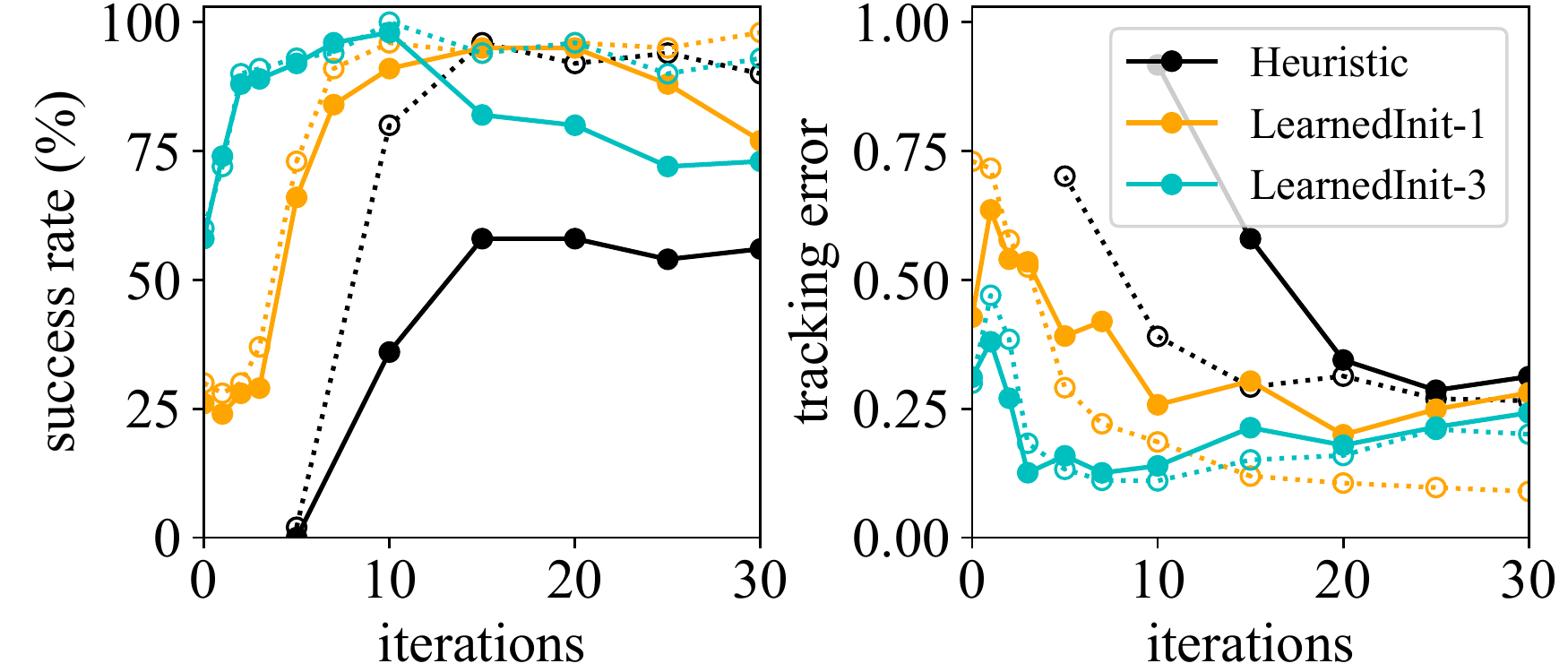}
    \caption{Performance of trajectories climbing a single pallet under whole-body control in simulation (100 tasks per data point). \learned{} produces solutions that are easier to track, reaching high success rates after minimal online refinement. Fixing phase durations at initialized values (dotted lines) improves optimization stability.}
	\label{fig:learning-sim}
	\vspace{-5mm}
\end{figure}

Fig.~\ref{fig:learning-sim} demonstrates that the outputs of \learned{} are executable more than half the time without any further optimization, and the success rate approaches 95\% within about 5~iterations of re-optimization. 
Similar results were achieved for all environments.
However, a higher number of iterations caused the optimization to unexpectedly diverge, resulting in a drop in success rates.
Shown by the dotted lines in Fig.~\ref{fig:learning-sim}, success rates remained high when a separate set of optimized trajectories, with phase durations fixed at their initialized values, were executed.
The observed behavior was caused by the extreme nonlinearity of the problem.

\subsection{Test on the real platform}
Table~\ref{tab:computation_time} summarizes the computation times obtained on the onboard locomotion computer (Intel i7-7500U).
Three variations of problem formulation have been tested: optimizing all variables, keeping only phase durations fixed, and keeping both phase durations and footstep locations fixed.
Not re-optimizing the phase durations gives a great improvement in the computation time per iteration while the success rate is similar, or better.
Additionally, the foot trajectories can be taken from the initialization (but eventually adapting the height to correspond to the terrain) and not re-optimized, but this results in a slight increase in the computation time.
It appears that it is simpler for the optimization to adapt the foot positions than the base trajectory since the base is coupled to, and constrained by, all the feet.

\begin{table}
	\centering
	\begin{tabular}{|l|c|c|c|c|}
		\hline
		Optimization  & \multicolumn{2}{c|}{Flat Ground (ms)} & \multicolumn{2}{c|}{Pallets (ms)}\cr
		Variables & 1st iter. & Mean & 1st iter. & Mean \cr
		\hline
		Full 				& 315 [304] 	& 156 [126] 	& 843 [700] 	& 417 [305] \cr
		No Phase 			& 210 [202] 	& 69 [60] 		& 385 [371] 	& 117 [108] \cr
		No Phase\&Feet 		& 282 [282] 	& 84 [74] 		& 534 [524] 	& 177 [145] \cr
		\hline
	\end{tabular}
	\caption{Computation time per iteration. The first iter. takes longer due to the solver initialization, subsequent iteration times are approx. constant. 
		The mean was computed for 100 iters. 
		The values w/o costs are shown in the square brackets.}
	\label{tab:computation_time}
	\vspace{-5mm}
\end{table}

While the robot's sensors could have been used to create a model of the terrain (using its Intel RealSense depth camera), the model of the environment was instead loaded from a virtual model.
This ensured repeatability and avoided limits in sensor field of view and the resolution.
For the tests using pallets, the robot's front camera was used to read an AprilTag which gave the position and orientation of the obstacle with respect to the robot.
The robot's onboard state estimator \cite{BloeschTwoStateImplicit2018} was used as state input; measureable estimator drift was present.
The full system---the generation of the initial guess, the optimization of the trajectory and its execution by the whole-body controller---ran onboard the robot's computer.

Fig.~\ref{fig:pallet_snpashots} shows the kinematic model and the real robot executing the optimized trajectory on the single pallet while Fig.~\ref{fig:experiment} shows the double pallet.
The results show that despite the errors in state and terrain estimation, the robot realized the trajectory to a high degree of accuracy and precision.

\section{Conclusion} \label{sec:conclu}
This work extended an optimization formulation for walking robot trajectories so that its solutions are not only feasible in theory, but can also be reliably executed on a real quadrupedal robot on a variety of terrains of increasing difficulty.
Strong nonconvexity and increased computational expense were offset by generating initial guesses from a neural network trained on filtered experiences gathered offline.
We will carry these findings onto our future work which aims to achieve online replanning of terrain-aware dynamic locomotion with a several-step horizon.
Future work will integrate environmental perception into the initialization map, more deliberately exploring the nonconvex solution space, and deploying the scheme in a receding-horizon manner for sustained mobility.

\bibliographystyle{IEEEtran}
\bibliography{library}
    
\end{document}